%% file: main.tex
\documentclass{article}

\usepackage[preprint]{neurips_2023}

\usepackage[utf8]{inputenc} 
\usepackage[T1]{fontenc}    
\usepackage{hyperref}       
\usepackage{url}            
\usepackage{booktabs}       
\usepackage{amsfonts}       
\usepackage{nicefrac}       
\usepackage{microtype}      
\usepackage{xcolor}         

\usepackage{amsmath}
\usepackage{multirow}
\usepackage{graphicx}
\usepackage{float}

\title{Adversarial Patch for 3D Local Feature Extractor}

\author{%
  Pao, Yu-Wen \\
  National Taiwan University\\
  Taipei, Taiwan \\
  \texttt{b09902016@csie.ntu.edu.tw}
  \And
  Li Chang Lai \\
  National Taiwan University\\
  Taipei, Taiwan \\
  \texttt{b09902135@csie.ntu.edu.tw}
  \And
  Lin, Hong-Yi \\
  National Taiwan University\\
  Taipei, Taiwan \\
  \texttt{b09902100@csie.ntu.edu.tw}
}

\begin{document}

\maketitle

\begin{abstract}
    \input{abstract}
\end{abstract}

\section{Introduction}\label{sec:introduction}
\input{intro}

\section{Related works}\label{sec:related-works}
\input{related_works}

\section{Methods}\label{sec:methods}
\input{method}

\section{Experimental results}\label{sec:experimental-results}
\input{result}

\section{Discussions}\label{sec:discussions}
\input{discussion}

\bibliographystyle{plain}
\bibliography{refs}

\section{Appendix}
\input{appendix}
\end{document}

%% file: abstract.tex
Local feature extractors are the cornerstone of many computer vision tasks. However, their vulnerability to adversarial attacks can significantly compromise their effectiveness. This paper discusses approaches to attack sophisticated local feature extraction algorithms and models to achieve two distinct goals: (1) forcing a match between originally non-matching image regions, and (2) preventing a match between originally matching regions. At the end of the paper, we discuss the performance and drawbacks of different patch generation methods.

%% file: intro.tex

Local feature extractors have become the backbone of many computer vision tasks that have revolutionized our world. Self-driving cars, for instance, rely heavily on accurate feature extraction to navigate safely. However, what if these powerful models misinterpret what they see?

This paper explores a specific adversarial attack that exploits how deep learning models interpret visual information. Generally, these models rely on local feature extractors to detect tiny snippets of an image, like edges or textures, to make sense of the bigger picture. This research paper examines how generating minor adjustments to an image can lead the model to misinterpret a scene.

Imagine a self-driving car encountering a stop sign. The car's computer vision model identifies the red octagon with local features. Our approach involves placing two small patches on the sign that appear different depending on the angle you look from. By confusing the local features, we hope to show how the model might misinterpret the entire scene, potentially with disastrous results.
Our implementation can be found at here\footnote{\href{https://github.com/paoyw/AdversarialPatch-LocalFeatureExtractor}{https://github.com/paoyw/AdversarialPatch-LocalFeatureExtractor}}

%% file: related_works.tex
\subsection{Local feature extraction}

The local feature extraction is to describe the image based on each local area of the image. The local feature extraction usually comes with two stages. The first stage, also known as feature detection, is to locate a set of points, objects, or regions in the images. The second stage is to create a descriptor for each feature point. In this work, we concentrate on SuperPoint\cite{detone2018superpoint}, a local feature extractor based on deep learning. The SuperPoint is a CNN-based model. The input will first passed into the encoder to encode a shared representation for the interest point decoder and the descriptor decoder. The interest point decoder can be seen as a classifier to find the position of the feature point for each non-overlapped $8\times 8$ region. The descriptor decoder gives the $256$ channel descriptions for each region.

\begin{figure}[ht]
    \centering
    \includegraphics[scale=0.2]{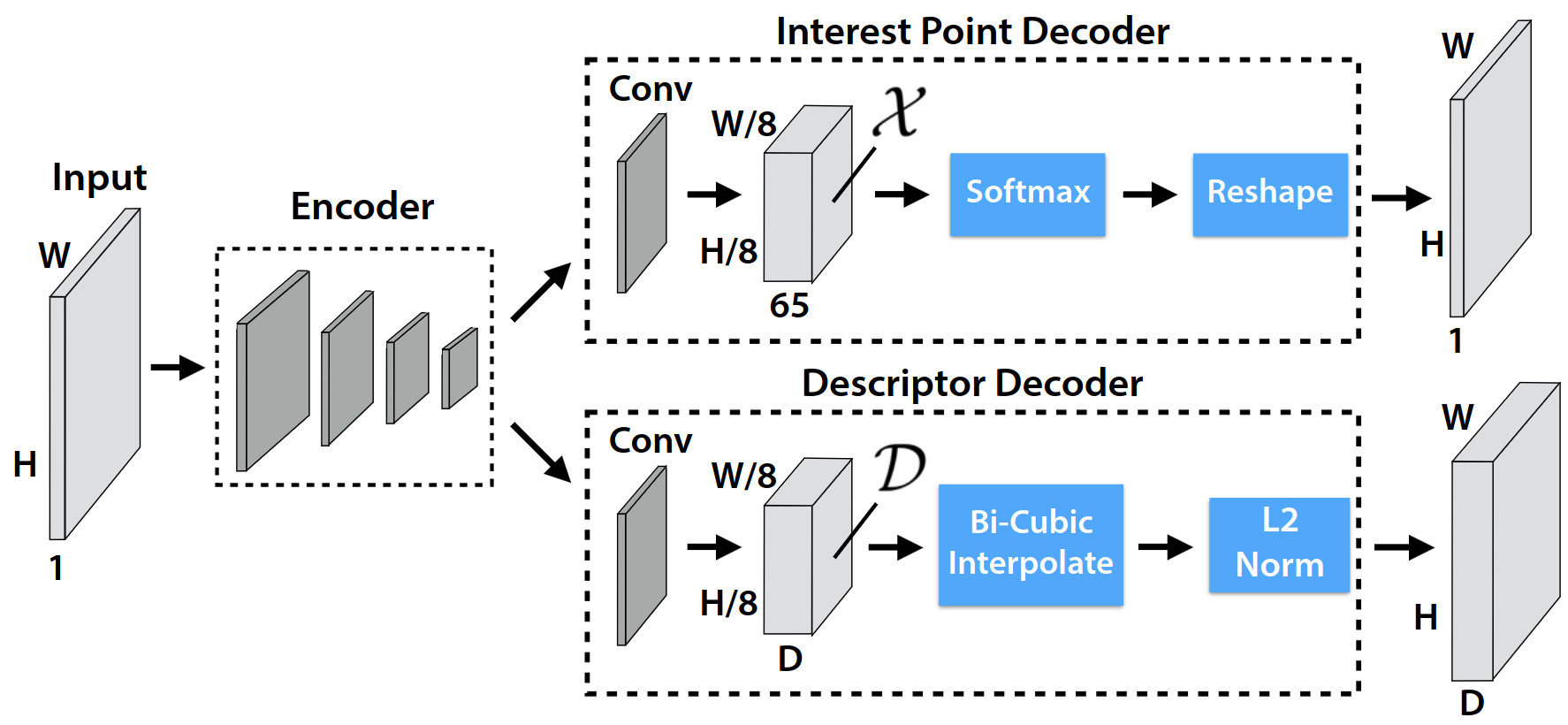}
    \caption{The model architecture of the SuperPoint\cite{detone2018superpoint}}
    \label{fig:superpoint}
\end{figure}

\subsection{Projective transformation}
Projective transformation\cite{hartley2003multiple}, also known as the homography, describes the change of the perceived object when the viewpoint changed by a $3\times3$ matrix, $H$, which is a homogeneous matrix.
$$
\begin{pmatrix}
x_{1}' \\
x_{2}' \\
x_{3}' \\
\end{pmatrix} = 
\begin{bmatrix}
    h_{11} & h_{12} & h_{13} \\
    h_{21} & h_{22} & h_{23} \\
    h_{31} & h_{32} & h_{33} \\
\end{bmatrix}
\begin{pmatrix}
x_{1} \\
x_{2} \\
x_{3} \\
\end{pmatrix}
$$
To be more specific, for a point, $(x, y)$, transform to a point $(x', y')$ when changing to the new viewpoint by applying a homography $H$.
It will be:
$$
x' = \frac{h_{11}x + h_{12} y + h_{13}}{h_{31}x + h_{32} y + h_{33}},
y' = \frac{h_{21}x + h_{22} y + h_{23}}{h_{31}x + h_{32} y + h_{33}}
$$

%% file: method.tex
\subsection{Overview}
There will be two adversarial patches in the same scene.
We denote the source patch as $P_{source}$. The other one, the target patch, is denoted as $P_{target}$. For the different viewpoints, the source view, $V_{source}$, and the target view, $V_{target}$, we want to increase the number of mismatches between the source patch at the source view and the target patch from the target view. The higher the mismatch rate is, the more likely it is to fail the downstream tasks. The proposed attack is composed of two parts. One is to generate an adversarial patch that the local feature extractor is sensitive to, while the other part is to determine the mask to which the adversarial patch will be applied.

\subsection{Adversarial patch generation}
The baseline adversarial patch is the chessboard pattern.
Due to the local feature extraction design, every junction point between four blocks on the chessboard should be identified as a local feature point.
What's more, the targeted local feature extractor, SuperPoint\cite{detone2018superpoint}, uses synthetic data similar to the chessboard as the input of the pre-training.
Hence, the SuperPoint is sensitive to chessboard patterns naturally.
We use $8*8$ size for each small cell in the chessboard pattern.

Besides of handcraft pattern, we want to generate a pattern that SuperPoint is sensitive to based on its model weights directly.
Inspired by FGSM\cite{goodfellow2014explaining} and PGD\cite{madry2017towards}, we create the adversarial patch, $x$, by multiple steps of gradient ascent by the following formula:
$$
x^t = x^{t-1} + \alpha \nabla_{x^{t-1}} L
$$
where $\alpha$ can be seen as the learning rate, $L$ is the loss function at $t$ step.
Since the interest point detector is a classifier, we can design two scenarios, one is the targeted class and the other is the untargeted class.
Their loss functions will respectively be:
$$
L_{ce}(\theta, x, y_{target}) \textit{ and } -L_{ce}(\theta, x, y_{dustbin})
$$
where $L_{ce}$ is the cross-entropy loss, $\theta$ is the model weight, $y_{target}$ can be any position in a $8\times 8$ patch, and $y_{dustbin}$ indicates the class that there's no local feature in the area.

Based on the early experimental results in \ref{sec:scale-invariant}, we found that the inconsistent size of the patch and the mask may cause a decrease in the performance.
Hence, we add augmentation like resizing and random cropping to increase the ability of the scale-invariant of the adversarial patch.

However, most of the performance of the chessboard pattern is better than the adversarial patch based on the experimental results \ref{sec:experimental-results}.
Hence, we try to directly inherit the performance of the chessboard and further boost the performance of the attack.
Instead of the random noisy or gray-scale image, we use the chessboard as the initial image for the optimization.
And then, apply the update with the augmentation.

\begin{figure}[ht]
    \centering
    \begin{minipage}{0.19\columnwidth}
    \centering
    \includegraphics[scale=0.55]{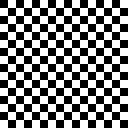}
    chessboard
    \end{minipage}
    \begin{minipage}{0.19\columnwidth}
    \centering
    \includegraphics[scale=0.55]{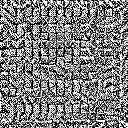}
    targeted adv.
    \end{minipage}
    \begin{minipage}{0.19\columnwidth}
    \centering
    \includegraphics[scale=0.55]{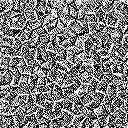}
    untargeted adv.
    \end{minipage}
    \begin{minipage}{0.19\columnwidth}
    \centering
    \includegraphics[scale=0.55]{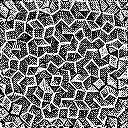}
    aug.
    \end{minipage}
    \begin{minipage}{0.19\columnwidth}
    \centering
    \includegraphics[scale=0.55]{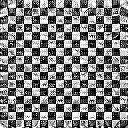}
    chess-init
    \end{minipage}
    \caption{The adversarial patches.}
    \label{fig:adv-patches}
\end{figure}

\subsection{Mask generation}

Mask generation determine the position and the shape, $P_{source}$ and $P_{target}$, that the adversarial patch will be filled in.
Since the intuition is to increase the similarity between $P_{source}$ at $V_{source}$, and $P_{target}$ at $V_{target}$, $P_{target}$ at $V_{source}$ should be similar to $P_{source}$ at $V_{srouce}$ after applied the homography transformation matrix, $H$, from $V_{source}$ to $V_{target}$.
Let's simply denote $P_{source}$ at $V_{source}$, as $P_{source}$, $P_{target}$ at $V_{source}$, as $P_{target}$, and $P_{target}$ at $V_{target}$, $P_{target}'$.
$$
 P_{source} \sim P_{target}'
$$
$$
 P_{source} \sim H P_{target}
$$
$$
 P_{source} = H H^{-1} P_{source}
$$
Hence, we can simply design $P_{target}$ as $H^{-1}P_{souce}$.
What's more, we can add some translations, which won't hurt the similarity between $P_{souce}$ and $P_{target}'$, to prevent overlapping or truncation by the image.

\begin{figure}[ht]
    \centering
    \includegraphics[scale=0.4]{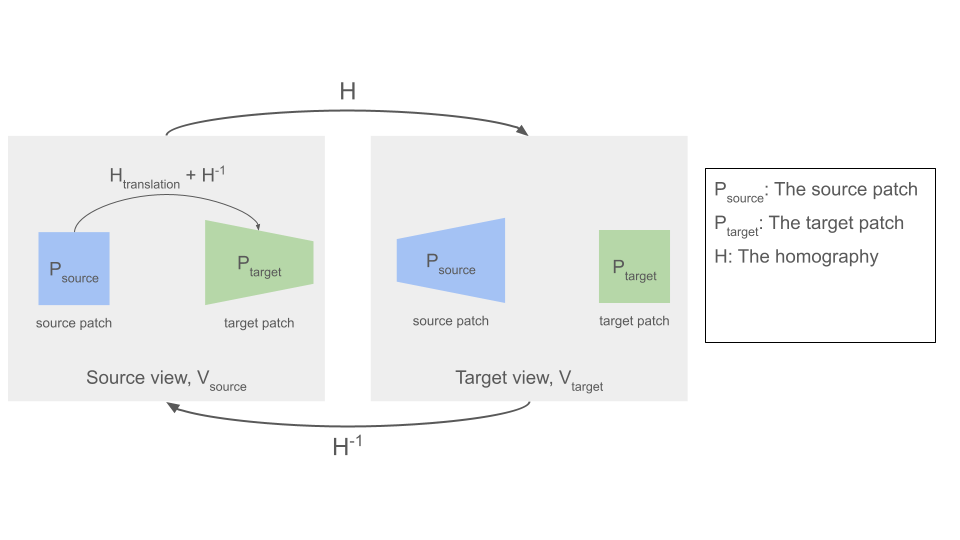}
    \caption{The generation of the masking for two patches}
    \label{fig:mask-gen}
\end{figure}

\subsection{Dataset}
We use HPatches\cite{balntas2017hpatches} as the dataset to evaluate the performance of the attack.
HPatches is composed of two parts.
The 59 patches are extracted from the image sequences with the viewpoint changes, and the other 57 patches are extracted from the sequence with the illumination changes.
We only take the 59 patches with the viewpoint changes for evaluation.

For each patch, there is one reference image and five compared images with the homography transformation matrix between them.
Then, we synthesize the adversarial patches on the images.
To fill the mask with the generated patches, we use backward warping with bi-linear interpolation.

In the targeted viewpoint settings, we compute the position and the shape of the mask for each pair of the reference image and the compared image.
In the untargeted one, we randomly select a compared image and compute the mask for each scene at the first step.
Then, we use the homography matrix provided by the dataset to compute the position of the mask from the previous step for the other viewpoints.
Hence, the mask of the same scene will be consistent in the 3D space.
The default setting is the targeted viewpoint.

\begin{figure}[ht]
    \centering
    \includegraphics[scale=0.6]{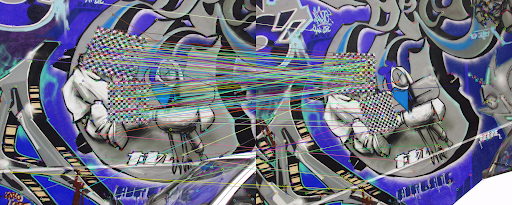}
    \caption{The visual result of the matching of a scene from two viewpoints}
    \label{fig:visual-result}
\end{figure}

\subsection{Metrics}
Without specification, we select the top-1000 points by k-NN matching.

\textbf{Source point ratio} means the number of the point detected in the source mask of the source view over the number of the point in the source view.

\textbf{True positive rate} means the number of the point  that is detected in the source mask of the target view over the number of the the number of the point detected in the source mask of the source view.

\textbf{False positive rate} means the number of the point  that is detected in the target mask of the target view over the number of the the number of the point detected in the source mask of the source view.

\textbf{Repeatability} evaluates that the same interest point should be detected for each scene.
First, use the ground truth homography to transform the interest points from the source view to the target view.
Then, take a pair of points from the source view and the target view that are close enough ($\epsilon =3$) to the same point.

\textbf{Homography estimation} can be viewed as a downstream task to evaluate the quality of the local feature points.
First, match the predicted local feature points point from two views by k-NN.
Then, use the RANSAC\cite{fischler1981random} to predict the transformation matrix.
Since directly comparing two homographies is not trivial, we utilize the four corners of the source view.
If the position of a corner is closed enough after applying the ground truth homography and the predicted homography, we take it as a correct point.

%% file: result.tex
\subsection{Targeted and untargeted viewpoint}\label{sec:targeted-untargeted-viewpoint}

In this experiment, we evaluate the performance of the three basic patches, chessboard pattern, targeted-class adversarial patch, and untargeted-class adversarial patch, under the targeted viewpoint and the untargeted viewpoint.
Table \ref{tab:targeted-untargeted} is the result.
From the targeted viewpoint, the untargeted-class adversarial patch successfully increases the source point ratio and the false positive rate.
However, the chessboard pattern outperforms it in the true positive rate and the homography estimation.

\begin{table}[ht]
    \centering
    \begin{tabular}{|c|c|c|c|c|c|c|c|c|}
    \hline
         \multirow{2}{*}{Viewpoint} & \multirow{2}{*}{Patch} & \multirow{2}{*}{SPR. $\uparrow$} & \multirow{2}{*}{TP. $\downarrow$} & \multirow{2}{*}{FP. $\uparrow$} & \multirow{2}{*}{Rep.} & \multicolumn{3}{c|}{Homography estimation $\downarrow$} \\
    \cline{7-9}
          &  &   &  &  &  & $\epsilon=1$ & $\epsilon=3$ & $\epsilon=5$\\
    \hline
        \multicolumn{6}{|c|}{benign} & 0.29 & 0.51 & 0.60 \\
    \hline
        \multirow{3}{*}{targeted} &  chessboard & 0.0605 & \textbf{0.1560} & 0.6371 & 0.3968 & \textbf{0.22} & \textbf{0.39} & \textbf{0.44} \\
        \cline{2-9}
         & targeted adv. patch & 0.0404 & 0.1700 & 0.5157 & 0.5074 & 0.30 & 0.51 & 0.58 \\
        \cline{2-9}
         & untargeted adv. patch & \textbf{0.1164} & 0.1989 & \textbf{0.7055} & 0.5289 & 0.29 & 0.48 & 0.56 \\
    \hline
    \hline
        \multirow{3}{*}{untargeted} &  chessboard & 0.0622 & \textbf{0.1969} & \textbf{0.4449} & 0.5656 & \textbf{0.25} & \textbf{0.42} & \textbf{0.50} \\
        \cline{2-9}
         & targeted adv. patch & 0.0522 & 0.2174 & 0.3313 & 0.5685 & 0.32 & 0.53 & 0.60 \\
        \cline{2-9}
         & untargeted adv. patch & \textbf{0.1485} & 0.3402 & 0.4394 & 0.5823 & 0.29 & 0.49 & 0.58\\
    \hline
    \end{tabular}
    \caption{The result of the targeted and untargeted view point}
    \label{tab:targeted-untargeted}
\end{table}

\subsection{The size of the mask}\label{sec:size-of-mask}
In this experiment, we evaluate the performance of the three basic patches, chessboard pattern, targeted-class adversarial patch, and untargeted-class adversarial patch, under three different sizes of the mask.
The generated patches are the same as the mask.
Table \ref{tab:mask-size} is the result.
We can see that the relative performance between different patches remains almost the same.
However, it is almost impossible to successfully attack on the homography estimation if the masking size is too small, due to the low source point ratio.

\begin{table}[ht]
    \centering
    \begin{tabular}{|c|c|c|c|c|c|c|c|c|}
    \hline
         \multirow{2}{*}{Size of mask} & \multirow{2}{*}{Patch} & \multirow{2}{*}{SPR. $\uparrow$} & \multirow{2}{*}{TP. $\downarrow$} & \multirow{2}{*}{FP. $\uparrow$} & \multirow{2}{*}{Rep.} & \multicolumn{3}{c|}{Homography estimation $\downarrow$} \\
    \cline{7-9}
          &  &   &  &  &  & $\epsilon=1$ & $\epsilon=3$ & $\epsilon=5$\\
    \hline
        \multicolumn{6}{|c|}{benign} & 0.29 & 0.51 & 0.60\\
    \hline
        \multirow{3}{*}{64} & chessboard & 0.0234 & \textbf{0.1396} & 0.6032 & 0.5781 & \textbf{0.29} &\textbf{0.50} & \textbf{0.57}\\
        \cline{2-9}
         & targeted adv. patch & 0.0087 & 0.2177 & 0.3064 & 0.5844 & 0.29 & 0.52 & 0.61 \\
        \cline{2-9}
        & untargeted adv. patch & \textbf{0.0566} & 0.1690 & \textbf{0.6665} & 0.5793 & 0.30 & 0.53 & 0.62\\
    \hline
    \hline
        \multirow{3}{*}{128} & chessboard & 0.0605 & \textbf{0.1560} & 0.6371 & 0.3968 & \textbf{0.22} & \textbf{0.39} & \textbf{0.44} \\
        \cline{2-9}
         & targeted adv. patch & 0.0404 & 0.1700 & 0.5157 & 0.5074 & 0.30 & 0.51 & 0.58 \\
        \cline{2-9}
        & untargeted adv. patch & \textbf{0.1164} & 0.1989 & \textbf{0.7055} & 0.5289 & 0.29 & 0.48 & 0.56 \\
    \hline
    \hline
        \multirow{3}{*}{256} & chessboard & 0.1841 & 0.2827 & 0.7086 & 0.3551 & \textbf{0.07} & \textbf{0.13} & \textbf{0.15}\\
        \cline{2-9}
         & targeted adv. patch & 0.1986 & \textbf{0.2778} & \textbf{0.7307} & 0.4886 & 0.30 & 0.51 & 0.59 \\
        \cline{2-9}
        & untargeted adv. patch & \textbf{0.2614} & 0.3274 & 0.7287 & 0.5221 & 0.28 & 0.46 & 0.52\\
    \hline
    \end{tabular}
    \caption{The result of the different size of mask.}
    \label{tab:mask-size}
\end{table}

\subsection{Scale-invariant}\label{sec:scale-invariant}
In this discussion, we want to test the scale-invariant of the adversarial patch.
In other words, will the inconsistent size of the patch and the mask affect the attack?
In the meantime, we introduce the augmentation and the initialization from the chessboard into the comparison.
Table \ref{tab:scale-invariant} is the result with $128$ as the size of the patch.
In the same size scenario, the chess-init patch has the highest performance overall, followed by the chessboard pattern.
When the patch size is slightly larger than that of the mask, the relative performance remains almost the same.
However, when the patch size is larger, the chessboard outperforms others once again.
Besides, the augmentation version of the patch is slightly better than the original untarget class version, but it brings lower performance on the homography estimation.

\begin{table}[ht]
    \centering
    \begin{tabular}{|c|c|c|c|c|c|c|c|c|}
    \hline
         \multirow{2}{*}{Size of patch} & \multirow{2}{*}{Patch} & \multirow{2}{*}{SPR. $\uparrow$} & \multirow{2}{*}{TP. $\downarrow$} & \multirow{2}{*}{FP. $\uparrow$} & \multirow{2}{*}{Rep.} & \multicolumn{3}{c|}{Homography estimation $\downarrow$} \\
    \cline{7-9}
          &  &   &  &  &  & $\epsilon=1$ & $\epsilon=3$ & $\epsilon=5$\\
    \hline
        \multicolumn{6}{|c|}{benign} & 0.29 & 0.51 & 0.60\\
    \hline
        \multirow{5}{*}{128} & chessboard & 0.0605 & \textbf{0.1560} & 0.6371 & 0.3968 & \textit{0.22} & \textit{0.39} & \textit{0.44} \\
        \cline{2-9}
         & targeted adv. patch & 0.0404 & \textit{0.1700} & 0.5157 & 0.5074 & 0.30 & 0.51 & 0.58 \\
        \cline{2-9}
        & untargeted adv. patch & 0.1164 & 0.1989 & 0.7055 & 0.5289 & 0.29 & 0.48 & 0.56 \\
        \cline{2-9}
        & aug. patch & \textit{0.1791} & 0.2464 & \textit{0.7360} & 0.6502 & 0.27 & 0.48 & 0.56 \\
        \cline{2-9}
        & chess-init patch &  \textbf{0.1968} & 0.1814 & \textbf{0.8250} & 0.4378 & \textbf{0.19} & \textbf{0.33} & \textbf{0.38} \\
    \hline
    \hline
        \multirow{5}{*}{100} & chessboard & \textit{0.1612} & \textbf{0.1520} & \textbf{0.8971} & 0.6274 & \textbf{0.24} & \textbf{0.40} & \textbf{0.46}\\
        \cline{2-9}
         & targeted adv. patch & 0.0393 & \textit{0.2263} & 0.6109 & 0.5788 & 0.30 & 0.53 & 0.61 \\
        \cline{2-9}
        & untargeted adv. patch & 0.0875 & 0.2836 & 0.6642 & 0.6034 & 0.31 & 0.54 & 0.62 \\
        \cline{2-9}
        & aug. patch & 0.1249 & 0.3067 & 0.6634 & 0.6555 & 0.33 & 0.55 & 0.63 \\
        \cline{2-9}
        & chess-init patch & \textbf{0.1653} & 0.2361 & \textit{0.7852} & 0.6443 & \textit{0.29} & \textit{0.48} & \textit{0.55} \\
    \hline
    \hline
        \multirow{5}{*}{150} & chessboard & 0.0185 & \textbf{0.1284} & 0.0816 & 0.5813 & \textit{0.25} & \textit{0.43} & \textit{0.50}\\
        \cline{2-9}
         & targeted adv. patch & 0.0331 & 0.1707 & 0.5216 & 0.5708 & 0.28 & 0.52 & 0.59 \\
        \cline{2-9}
        & untargeted adv. patch & 0.1184 & 0.1916 & 0.7421 & 0.5724 & 0.27 & 0.49 & 0.58 \\
        \cline{2-9}
        & aug. patch & \textbf{0.1781} & 0.2245 & \textit{0.7640} & 0.6056 & 0.26 & 0.46 & 0.51 \\
        \cline{2-9}
        & chess-init patch & \textit{0.1307} & \textit{0.1457} & \textbf{0.847}1 & 0.5341 & \textbf{0.21} & \textbf{0.38} & \textbf{0.45} \\
    \hline
    \hline
        \multirow{3}{*}{64} & chessboard & \textbf{0.0790} & \textbf{0.1363} & \textbf{0.8783} & 0.6562 & \textbf{0.29} & \textbf{0.51} & \textbf{0.59}\\
        \cline{2-9}
         & targeted adv. patch & 0.0081 & 0.4059 & 0.1404 & 0.5925 & \textit{0.30} & \textit{0.52} & \textit{0.62}\\
        \cline{2-9}
        & untargeted adv. patch & \textit{0.0207} & \textit{0.3076} &  \textit{0.4090} & 0.5903 & \textit{0.30} & 0.53 & \textit{0.62} \\
    \hline
    \hline
        \multirow{3}{*}{256} & chessboard & 0.0034 & \textbf{0.1864} & \textit{0.0706} & 0.5874 & \textbf{0.29} & 0.51 & 0.60 \\
        \cline{2-9}
         & targeted adv. patch & \textit{0.0079} & 0.2812 & 0.0655 & 0.5889 & 0.31 & 0.51 & 0.60 \\
        \cline{2-9}
        & untargeted adv. patch & \textbf{0.0096} & \textit{0.2237} & \textbf{0.0924} & 0.5849 & \textit{0.30} & 0.51 & 0.60 \\
    \hline
    \end{tabular}
    \caption{The result of the scale-invariant at small scale and large scale}
    \label{tab:scale-invariant}
\end{table}

\subsection{Transferability}\label{sec:transferability}
In this section, we evaluate the transferability to other local feature extractors of our attack.
We evaluate our attack on SIFT\cite{lowe2004distinctive} and SuperPoint\cite{detone2018superpoint}.
Table \ref{tab:transferability} is the result.
We only focus on two patches, the chessboard pattern and chess-init, based on the performance of the previous results.
We can see that these two patterns can successfully attack the SIFT as well.
However, the performance is not that well against SuperPoint.

\begin{table}[ht]
    \centering
    \begin{tabular}{|c|c|c|c|c|c|c|c|}
    \hline
         \multirow{2}{*}{Local feature extractor} & \multirow{2}{*}{Patch} & \multirow{2}{*}{SPR. $\uparrow$} & \multirow{2}{*}{TP. $\downarrow$} & \multirow{2}{*}{FP. $\uparrow$} & \multicolumn{3}{c|}{Homography estimation $\downarrow$} \\
    \cline{7-8}
          &  &   &  &  &  $\epsilon=1$ & $\epsilon=3$ & $\epsilon=5$\\
    \hline
        \multirow{4}{*}{SuperPoint\cite{detone2018superpoint}} & \multicolumn{4}{|c|}{benign} & 0.29 & 0.51 & 0.60\\
        \cline{2-8}
         & chessboard & 0.0605 & \textbf{0.1560} & 0.6371 & 0.22 & 0.39 & 0.44  \\
        \cline{2-8}
        & chess-init patch & \textbf{0.1968} & 0.1814 & \textbf{0.8250} & \textbf{0.19} & \textbf{0.33} & \textbf{0.38}\\
    \hline
        \multirow{4}{*}{SIFT\cite{lowe2004distinctive}} & \multicolumn{4}{|c|}{benign} & 0.39 & 0.59 & 0.66 \\
        \cline{2-8}
        & chessboard & \textbf{0.0810} & 0.4281 & 0.5888 & \textbf{0.37} & \textbf{0.54} & \textbf{0.59} \\
        \cline{2-8}
        & chess-init patch & 0.0235 & \textbf{0.1827} & \textbf{0.7490} & 0.37 & 0.55 & 0.59 \\
    \hline
    \end{tabular}
    \caption{The result of the transferability}
    \label{tab:transferability}
\end{table}

%% file: discussion.tex
To the best of our knowledge, we are the first to propose a patch-based adversarial attack against SuperPoint\cite{detone2018superpoint} even local feature extraction.
We successfully perform the attack on a well-known local feature extraction dataset, HPatches\cite{balntas2017hpatches} by synthesizing the adversarial patches.
Although we have shown some vulnerabilities of the local feature extraction and proposed a simple yet effective method to attack it, there is still a lot more to explore.
One of the possible directions is to design stronger patterns, which have a higher scale-invariant, smaller size of the mask.
To a certain degree, changing the two patches scenario to one patch only.

What's more, though the feature matching in our evaluation is kNN with RANSAC, there have been many works of the deep-learning-based local feature matching, like SuperGlue\cite{sarlin2020superglue} and LightGlue\cite{lindenberger2023lightglue}.
Designing an attack against both deep-learning-based local feature extraction and matching may be challenging and delicate work.

From the perspective of defenses, there have been some works \cite{amerini2011sift} \cite{huang2008detection} to detect copy-move forgery, which our attack can be somehow classified as.
We leave the debate between attacks and defenses of the adversarial attack against the local feature extraction as the future works.

Overall, we hope that this work provides a new perspective on the security of local feature extraction.
And we are looking forward to the growth of this topic.

%% file: appendix.tex
\subsection{Visual results of the scale-invariant experiment}

\begin{figure}[H]
    \centering
        \includegraphics[scale=0.15]{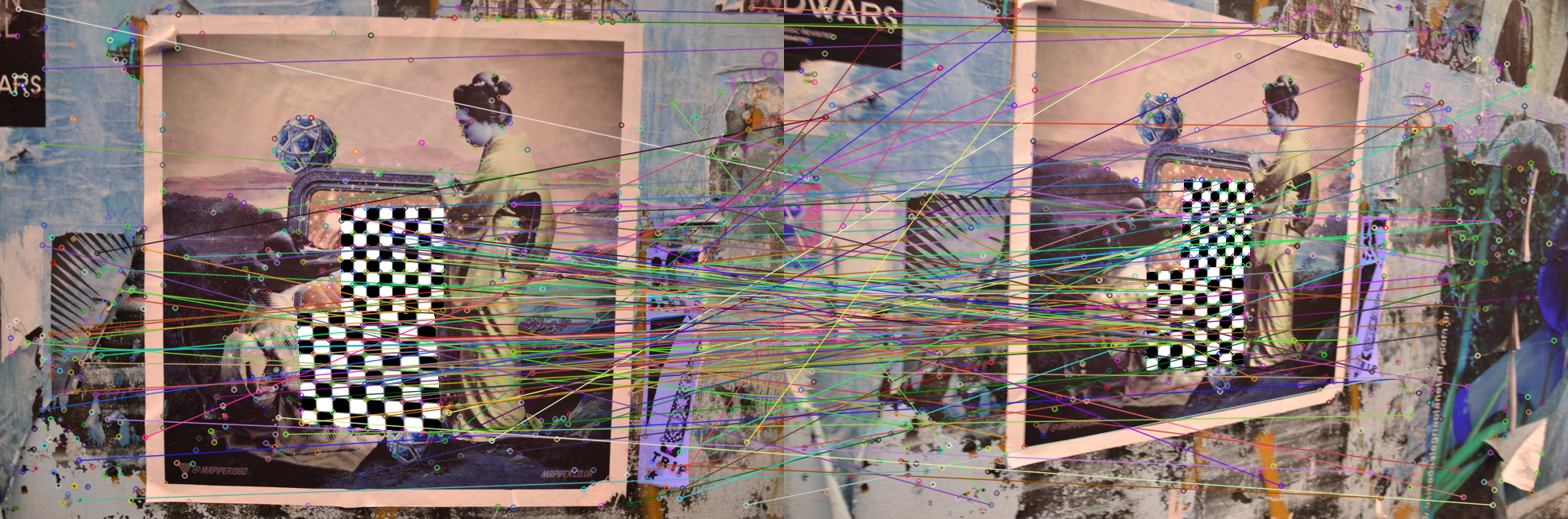} \\ patch size 64 \\
        \includegraphics[scale=0.15]{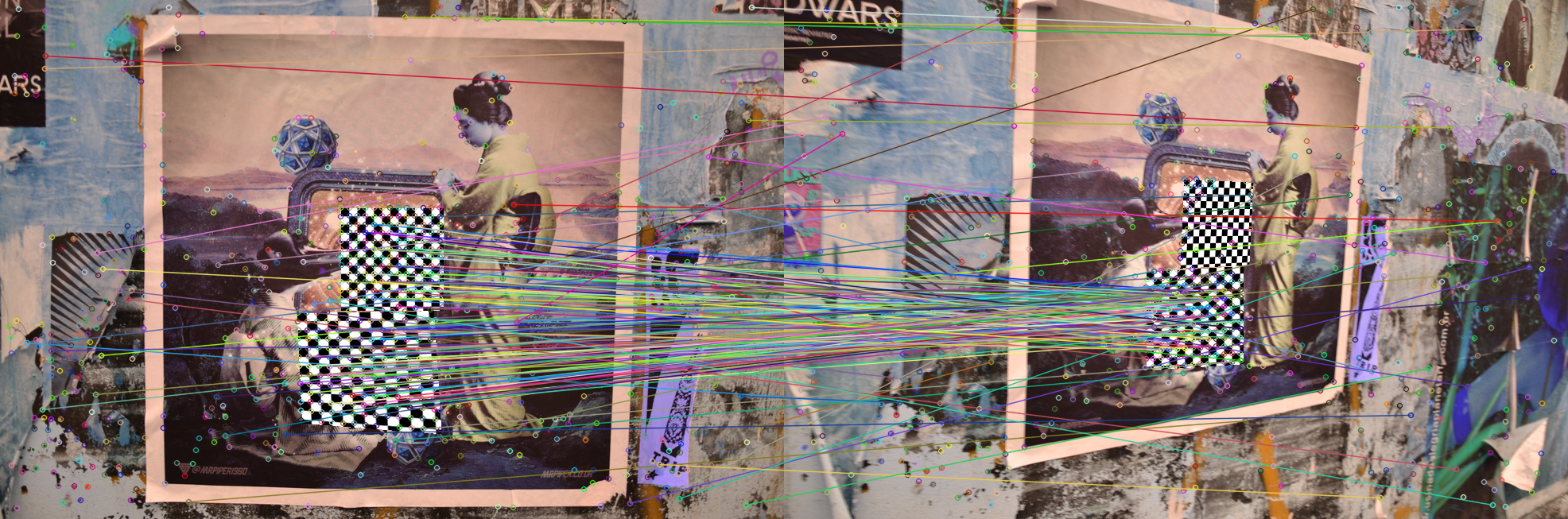} \\ patch size 100 \\
        \includegraphics[scale=0.15]{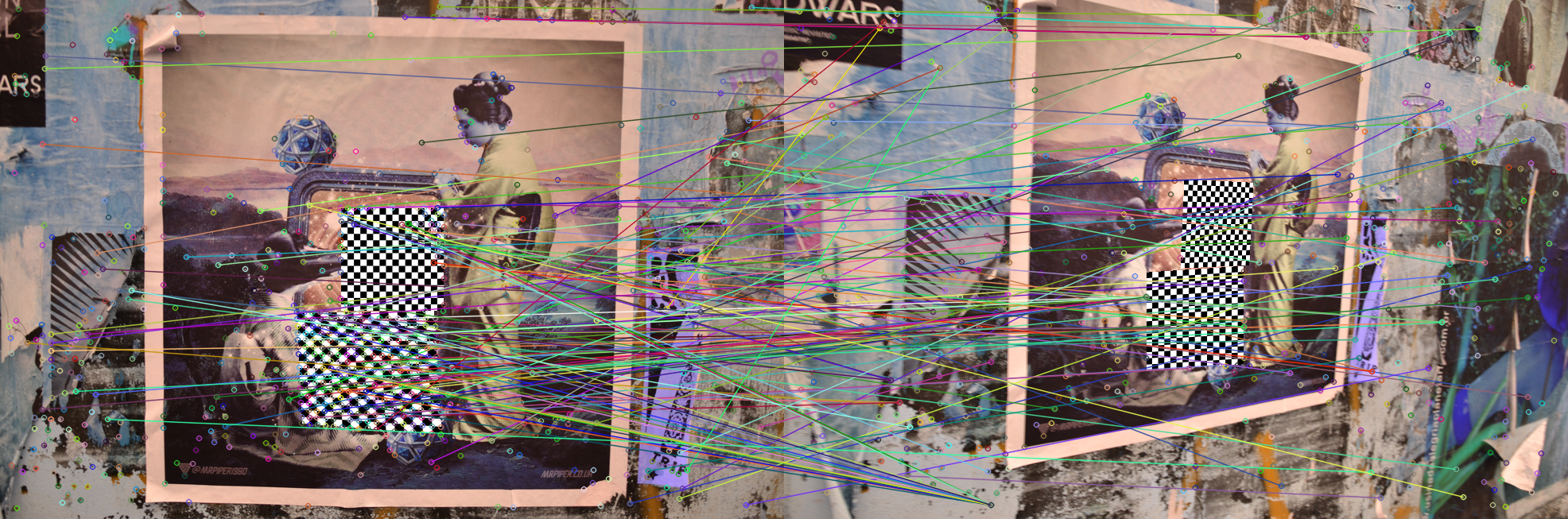} \\ patch size 128 \\
        \includegraphics[scale=0.15]{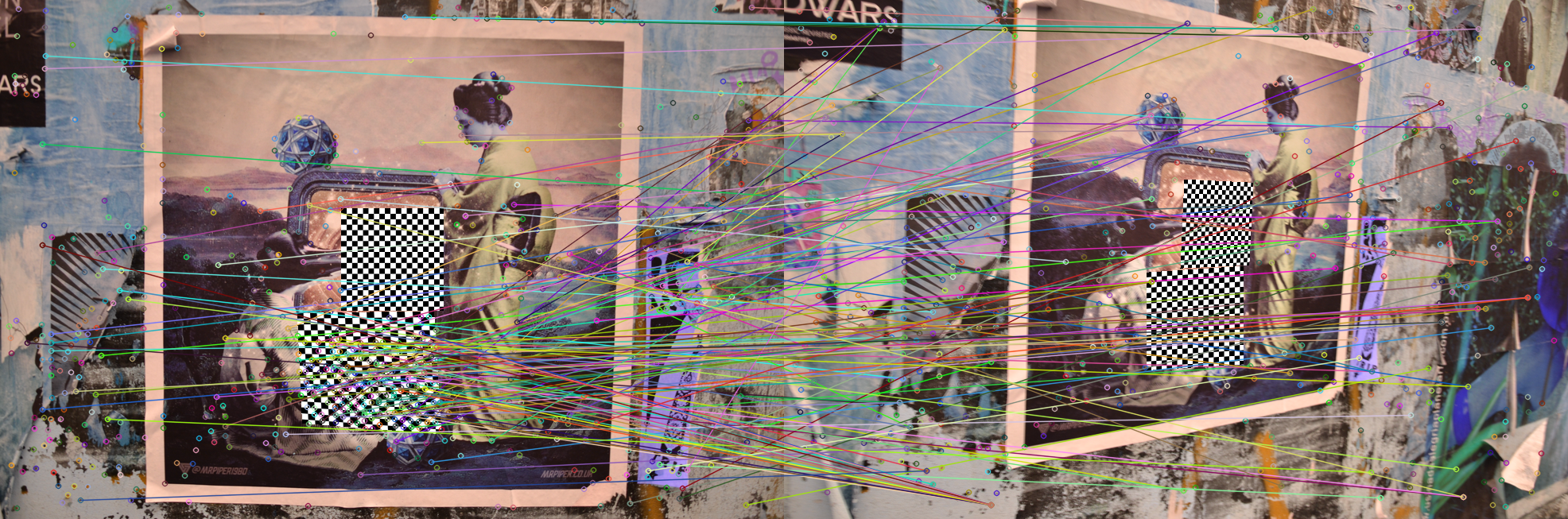} \\ patch size 150 \\
        \includegraphics[scale=0.15]{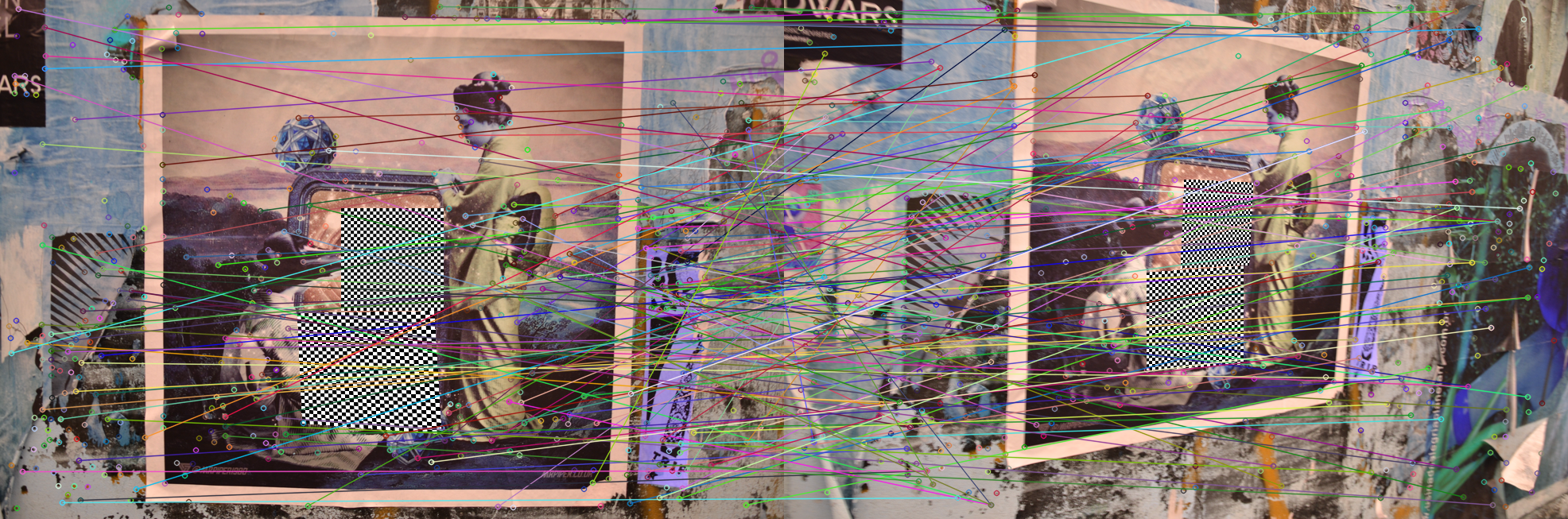} \\ patch size 256 \\
    \caption{The visual results of the scale-invariant experiment with top-150 matching points}
    \label{fig:scale-invariant-visual}
\end{figure}

\subsection{Visual result of the different size of the mask experiment}
\begin{figure}[H]
    \centering
        \includegraphics[scale=0.125]{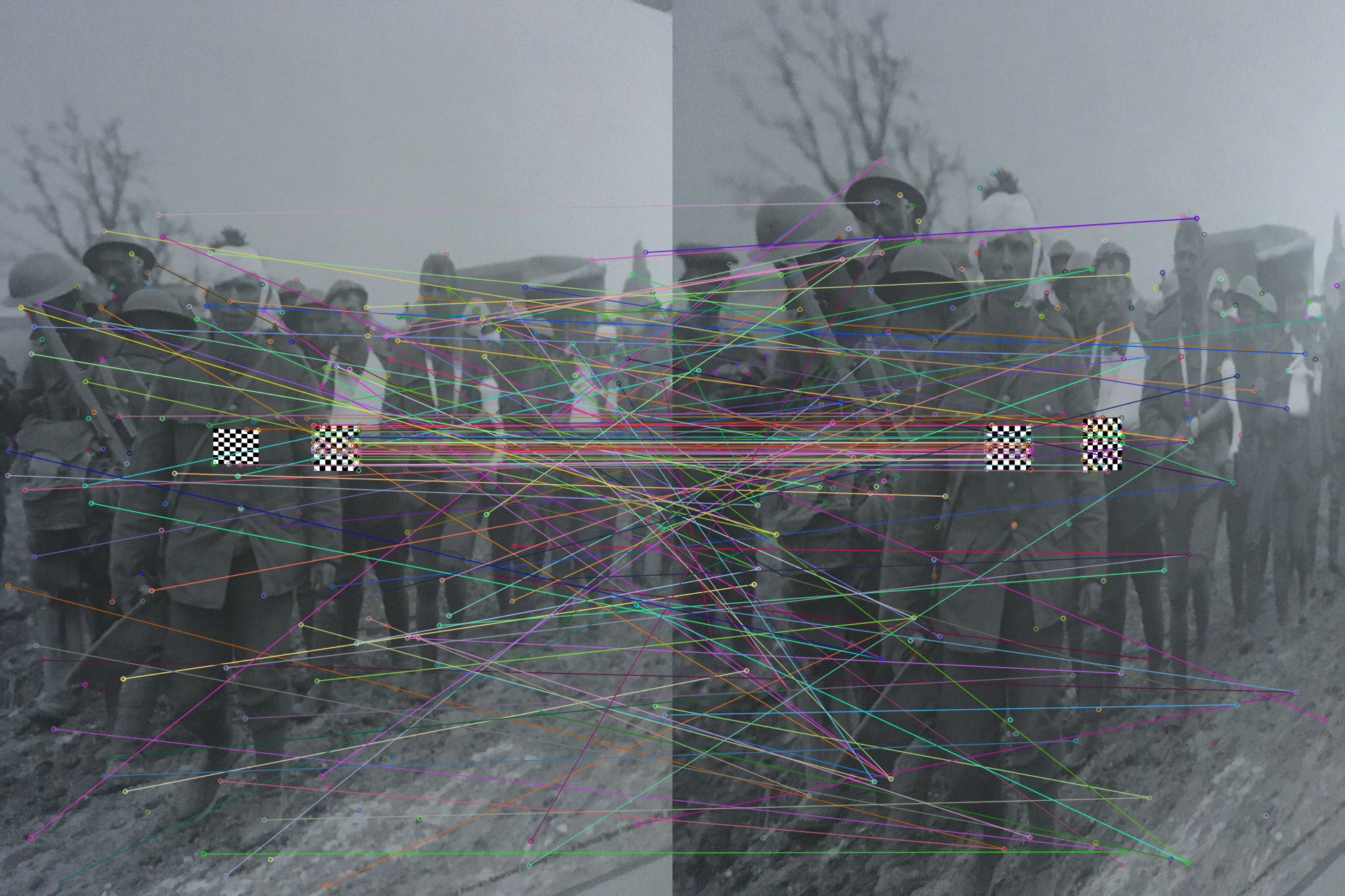} \\ mask size 64 \\
        \includegraphics[scale=0.125]{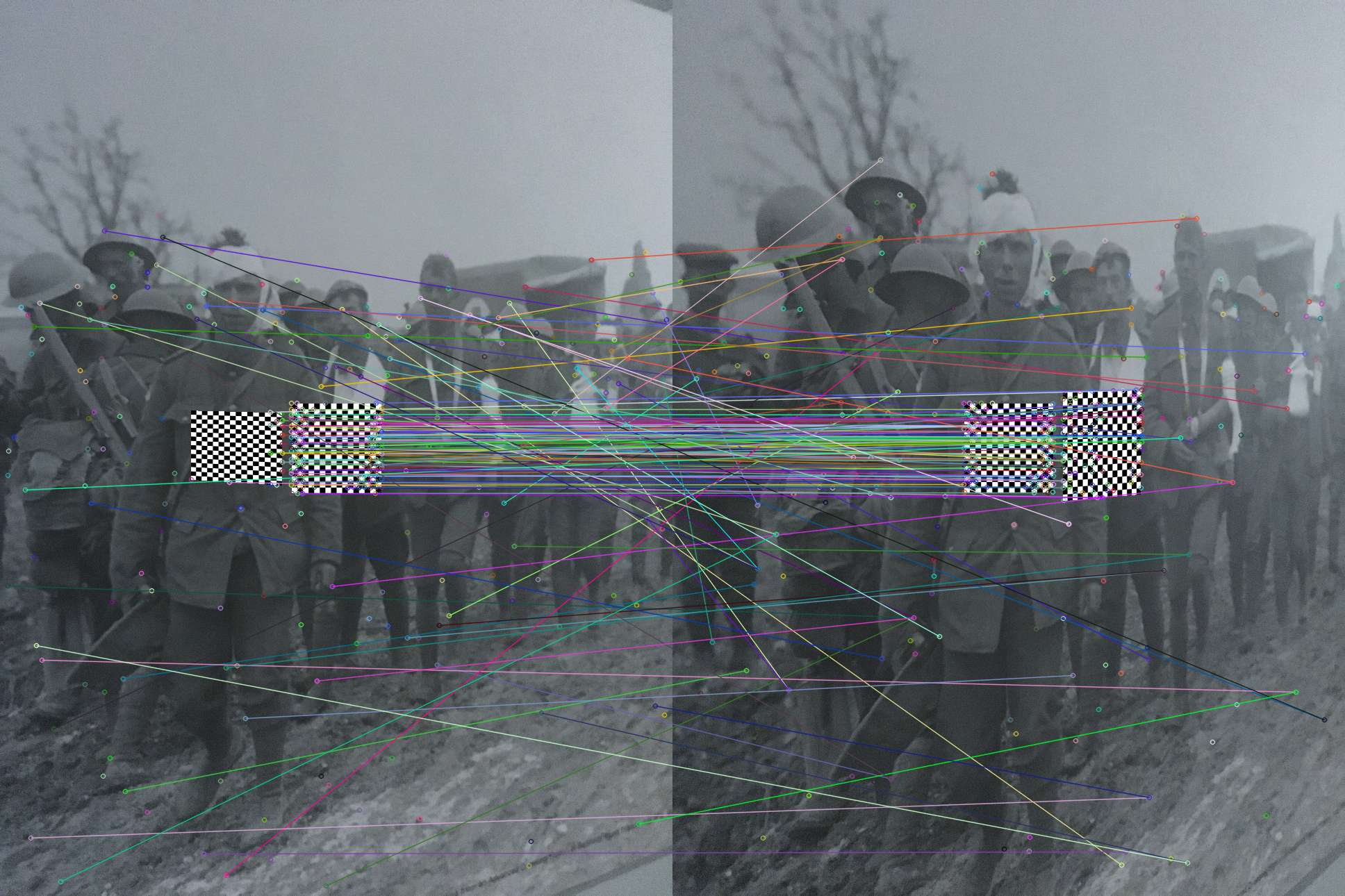} \\ mask size 128 \\
        \includegraphics[scale=0.125]{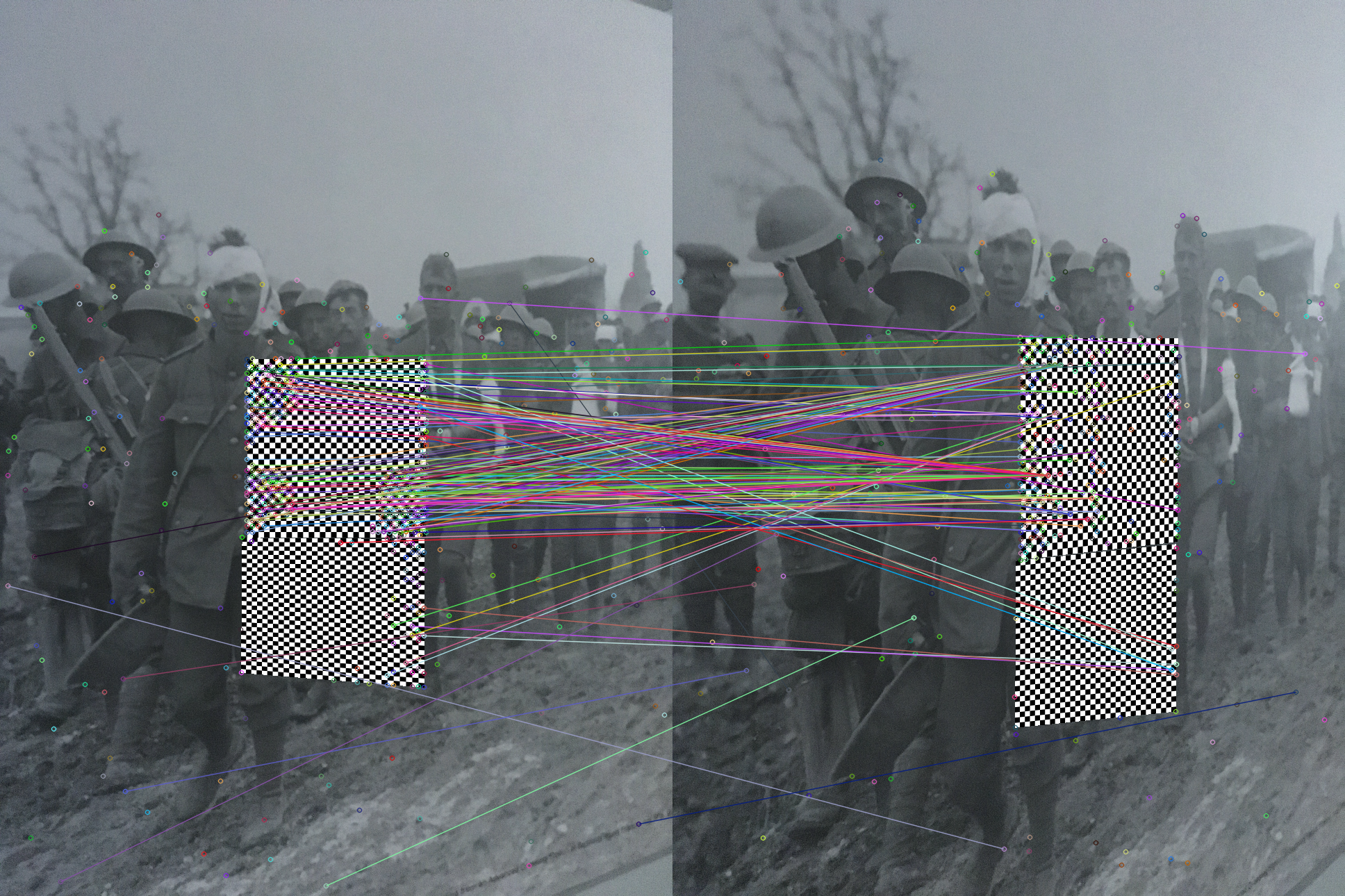} \\ mask size 256 \\
    \caption{The visual result of the different size of the mask with top-150 matching points}
    \label{fig:mask-size-visual}
\end{figure}

\subsection{Visual result of the transfer-ability experiment}
\begin{figure}[H]
    \centering
        \includegraphics[scale=0.25]{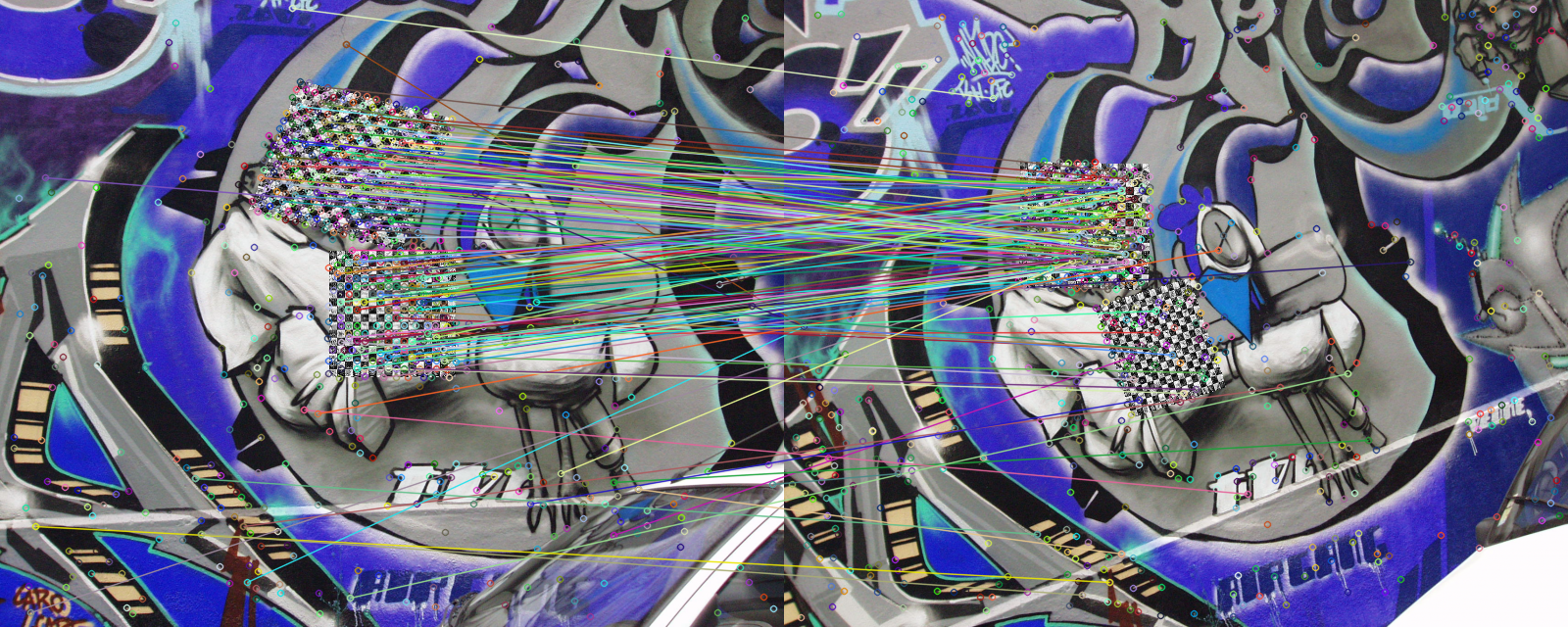} \\ SuperPoint with chess-init patch \\
        \includegraphics[scale=0.25]{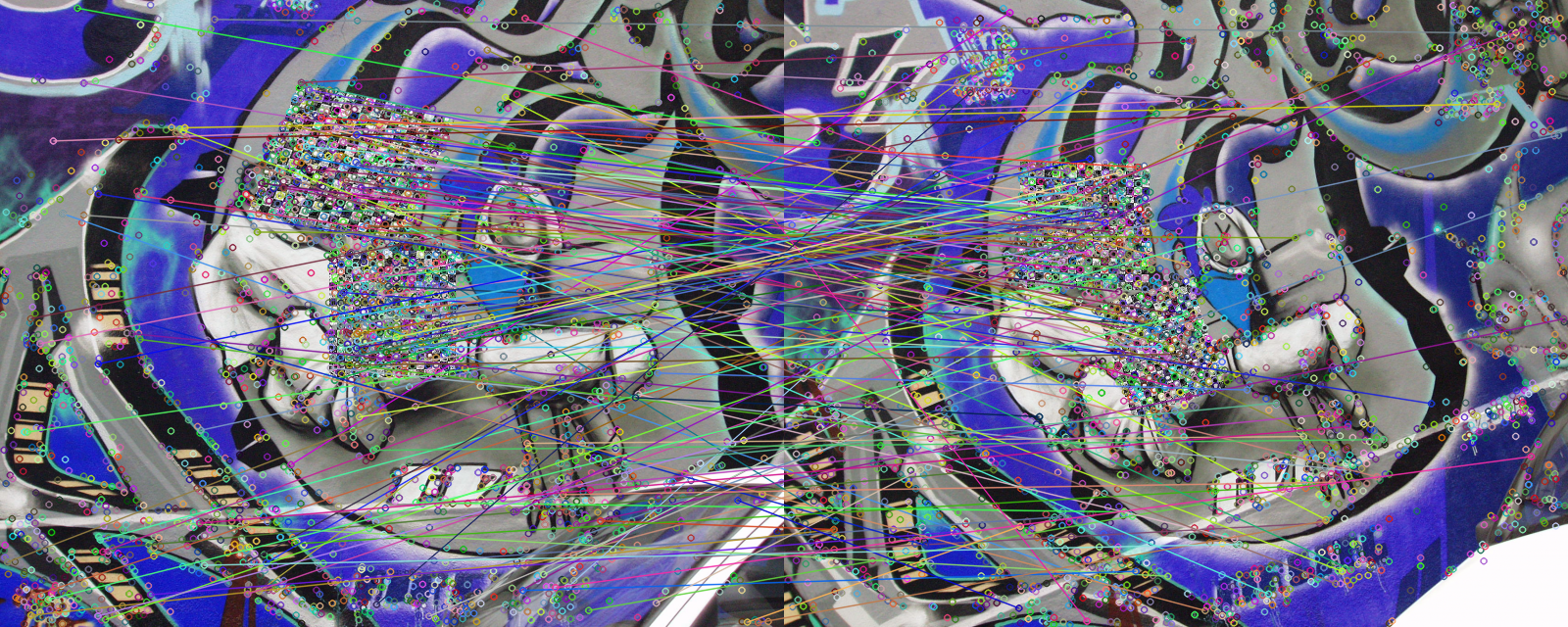} \\ SIFT with chess-init patch \\
    \caption{The visual result of the transfer-ability experiment with top-150 matching points}
    \label{fig:transferability-visual}
\end{figure}